\documentclass[letterpaper, 10 pt, conference]{ieeeconf}
\IEEEoverridecommandlockouts
\usepackage{graphics} 
\usepackage{epsfig} 
\usepackage{mathptmx} 
\usepackage{times} 
\usepackage{amsmath} 
\usepackage{amssymb}  
\usepackage{subcaption}
\usepackage{simplemacros}
\usepackage{hyperref}
\usepackage{xfrac}
\usepackage{algorithm}
\usepackage{algorithmic}
\usepackage{tikz}
\usepackage{titlesec}
\usepackage{placeins}

\usetikzlibrary{positioning,shapes}

\titlespacing\subsection{0pt}{4pt plus 4pt minus 2pt}{0pt plus 2pt minus 2pt}

\title{\LARGE \bf
Datasets, Models, and Algorithms for Multi-Sensor, Multi-Agent Autonomy Using \texttt{AVstack}
}

\author{Spencer Hallyburton and Miroslav Pajic
\thanks{*This work is sponsored in part by the ONR N00014-23-1-2206  and AFOSR FA9550-19-1-0169 awards, NSF  CNS-1652544, and the National AI Institute for Edge Computing Leveraging Next Generation Wireless Networks, Grant CNS-2112562.}
\thanks{The authors are with the Department of Electrical and Computer
    Engineering, Duke University, Durham, NC 27708 USA (e-mail:
    \{spencer.hallyburton, miroslav.pajic\}@duke.edu).}
}

\begin{document}

\maketitle

\thispagestyle{plain}
\pagestyle{plain}

\begin{abstract}

Recent advancements in assured autonomy have brought autonomous vehicles (AVs) closer to fruition. Despite strong evidence that multi-sensor, multi-agent (\msma) systems can yield substantial improvements in the safety and security of AVs, there exists no unified framework for developing and testing representative \msma\ configurations. Using the recently-released autonomy platform, \avstack, this work proposes a \emph{new framework for datasets, models, and algorithms} in \msma\ autonomy. 
Instead of releasing a single dataset, we deploy a \emph{dataset generation pipeline} capable of generating unlimited volumes of ground-truth-labeled \msma\ perception data. The data derive from cameras (semantic segmentation, RGB, depth), \lidar, and \radar\, and are sourced from ground-vehicles and, for the first time, infrastructure platforms. 
Pipelining generating labeled \msma\ data along with \avstack's third-party integrations defines a \emph{model training framework} that allows training multi-sensor perception for vehicle and infrastructure applications. We provide the framework and pretrained models open-source.
Finally, the dataset and model training pipelines culminate in insightful multi-agent case studies. While previous works used specific ego-centric multi-agent designs, our framework considers the collaborative autonomy space as a network of noisy, time-correlated sensors. Within this environment, we quantify the impact of the network topology and data fusion pipeline on an agent's situational awareness. 

\end{abstract}
\section{Introduction}

Unavoidable occlusion, sensor degradation due to adverse environmental conditions, and adversarial actors threaten the prospect of safely deploying autonomous vehicles (AVs) in the real world. An important avenue for overcoming these challenges is to pursue multi-sensor, multi-agent (\msma) collaboration. Multi-sensor systems leverage multiple modalities of perceptive elements within a platform including cameras, \lidar, and \radar. Multi-agent systems share information with external platforms via vehicle-to-vehicle (V2V) and vehicle-to-infrastructure channels. Both multi-sensor and multi-agent concepts are essential to the assuredness of perception, planning, and control. However, multi-agent research is immature and under-supported in the community. 

Recently, datasets have emerged to support multi-sensor AV development. The nuScenes~\cite{2020nuscenesdataset} and Waymo~\cite{2020waymodataset} datasets provide large numbers of scenes with multiple ego-centric sensors of multiple modalities including cameras, \lidar, and \radar. These datasets have become the gold-standard for AV research and have inspired many high-performing multi-sensor perception algorithms (e.g.,~\cite{zhuang2021perception, harley2023simple, fadadu2022multi}).

On the other hand, progress in the multi-agent domain has left much to be desired. Transportation experts believe development of V2V and V2I collaboration is essential for safely deploying AVs in the real world~\cite{nhtsav2v}. Unfortunately, manufacturers are isolated in developing their individual self-driving cars while government entities still work out how to contract the work for multi-agent collaboration. This large rift has contributed to the slowdown of self-driving vehicle innovation. As such, we need high-quality datasets to advance collaborative autonomy. 

Recently, the \emph{Open Dataset for Perception with V2V communication} (OPV2V)~\cite{2022openv2v} and research on connected AVs (CAV)~\cite{2022cooperative} provide mostly \lidar\ data from multiple ground-based agents. These provide a baseline for considering simple collaborative algorithms such as vehicle platooning (see~\cite{bergenhem2012platooning}). There, secondary agents send perception data to the ego vehicle. Upon receipt, the ego vehicle collates data and runs perception over the joint data. Unfortunately, both datasets provide simplified scenarios and neglect to consider representative conditions such as noisy, time-correlated infrastructure sensing. Moreover, neither dataset provides a \emph{general} framework for training perception models; both datasets instead provide limited, specialized perception models that can only be trained on their datasets.

The lack of a unified framework for pursuing multi-sensor, multi-agent analysis of perception systems in AVs is inhibiting progress in deploying assured AVs in challenging environments. Providing a new dataset \emph{alone}, however, would only be an \emph{incremental} improvement over the existing state of the art. Instead, in this work, \textbf{we establish a new \emph{framework} for multi-sensor, multi-agent autonomy.} This framework is made possible by the following contributions.

\vspace{4pt}
\noindent \textbf{Contribution 1: \msma\ dataset generation pipeline.} 
Prior datasets either derive from a physical environment or use a simulator to generate a limited dataset. Rather than propose a singular dataset for \msma\ research, we build a \emph{dataset-generation pipeline} using \avstack~\cite{avstack,avstack-demo} and \carla\ simulator~\cite{2017carla} that allows for creation of unlimited volumes of \msma\ data. Using \avstack, the interface to our datasets is standardized and interchangeable with existing benchmarks (e.g.,~KITTI, nuScenes, Waymo). To demonstrate the power of this new framework, we generate over 1~TB of \emph{open-source} ground-truth-labeled camera, \lidar, and \radar\ data using multiple vehicles and infrastructure platforms in only a few hours of real-time.

\vspace{4pt}
\noindent \textbf{Contribution 2: \msma\ model training framework.} 
We bridge the gap between \msma\ research and the growing collection of \avstack's third-party integrations. An important advantage of our framework is that a freshly-generated dataset is immediately compatible with \emph{dozens} of supported models for camera-based and LiDAR-based detection, classification, depth estimation, and semantic segmentation models. To showcase the power of this model training framework, we generate an open dataset and leverage \avstack's integration with OpenMMLab~\cite{chen2019mmdetection} to train the first infrastructure-based models for camera and \lidar\ perception to go along with high-performing vehicle-centric perception. We illustrate the importance of training application-specific models by showing that, in an infrastructure setting, camera-based perception trained on the \emph{ego vehicle} alone severely underperforms against the same model trained on our infrastructure data.

\vspace{4pt}
\noindent \textbf{Contribution 3: Multi-agent algorithms and case studies.}\\
Finally, we complete the multi-agent evaluation lifecycle by using the newly-generated datasets and newly-trained models in the design, implementation, test, and analysis (DITA) process. Prior multi-agent datasets consider a design where all non-ego agents communicate only with the ego agent~\cite{2022openv2v,2022cooperative}. A more likely scenario is that infrastructure and vehicle agents will compose a \emph{sensor network}. We use our datasets and models to spin up case studies considering practical multi-agent designs and network topologies. Under these ego and network models, our case studies find that performing naive Bayesian fusion within a noisy, time-correlated network underperforms distributed data fusion algorithms such as track fusion with covariance intersection~\cite{2017ddfwithCI}. 

In summary, we propose a new framework for multi-sensor, multi-agent autonomy. The framework is built on \avstack's~\cite{avstack,avstack-demo} integrations with the \carla\ simulator~\cite{2017carla} and third-party extensions such as OpenMMLab's perception suite~\cite{chen2019mmdetection}. This framework is made possible by the following innovations:
\begin{itemize}
    \item A dataset-generation pipeline for creating ground-truth-labeled multi-sensor, multi-agent datasets.
    \item The integration of \msma\ datasets with \emph{application-specific} object detection, classification, and segmentation perception models. 
    \item A DITA framework for novel multi-agent testing and evaluations in complex \msma\ sensor networks.
\end{itemize}
\section{Multi-Sensor, Multi-Agent Dataset Pipeline} \label{sec:2-dataset}

Early datasets in computer vision (CV) such as COCO~\cite{coco2014microsoft} and PASCAL~\cite{2010pascalvoc} focused on single-sensor, agent-agnostic snapshots to train e.g., perception models with supervised labels. Since the early CV works, domain-specific datasets built on longitudinal scenarios such as KITTI~\cite{2013kittidataset} and nuScenes~\cite{2020nuscenesdataset} have arisen to guide testing and model development of AVs. These datasets have contributed substantially to the advancement of perception.

While the prior datasets were important stepping stones in AV research, they have limitations. First, due to the large engineering effort required to capture data in the real-world, many are limited and fixed in size. Second, the available sensors are fixed. Third, the quantity/quality of usable data are limited due to the challenging task of labeling data. For example, only 15\% of the Cityscapes~\cite{cordts2016cityscapes} dataset is labeled for supervised learning. Fourth, in the physical world, the behavior and configuration of other agents is stochastic. This makes it impossible to ensure capturing scenarios of interesting data. Finally, to the best of our knowledge, no publicly available datasets make use of infrastructure sensors for collaborative autonomy; this shortcoming is severely limiting research in \msma\ autonomy.

We describe here our innovations that allow \avstack\ to be used as a new framework for generating research-quality \msma\ datasets from the \carla\ simulator. 

\subsection{Multi-Modal Sensor Compatibility}

Recently, researchers have proposed datasets for multi-agent, collaborative autonomy~\cite{2022openv2v,2022cooperative} from simulators. Existing multi-agent datasets consider collaborative sensing from other mobile agents. However, until now, no dataset considers the existence of \textit{infrastructure sensing}.
\begin{figure*}[t!]
    \centering
    \includegraphics[width=\linewidth]{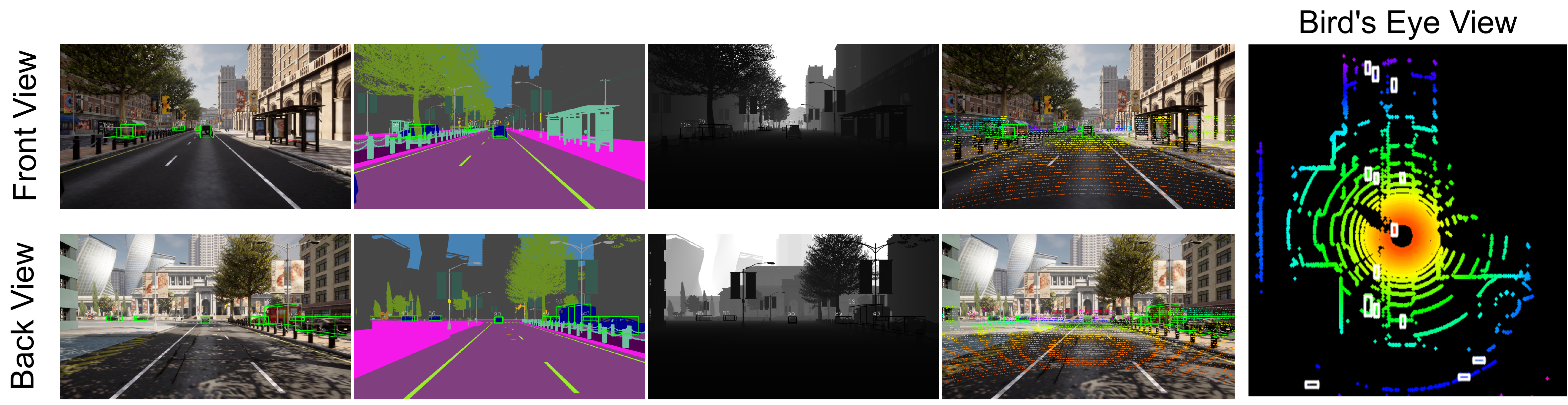}
    \caption{Our multi-sensor dataset provides an extensive suite of sensor data. Each data instance has ground-truth object labels and consistent identification between sensing modalities over time. (left-to-right) RGB, semantic segmentation, depth, and LiDAR projected onto camera; (top row) front-facing, (bottom row) rear-facing. (far right) Bird's eye view of LiDAR.}
    \label{fig:dataset-sensor-visualization}
\end{figure*}

To pipeline the dataset generation process, we employ the \carla\ simulator as our dataset generation engine. With the help of \avstack's expanded \carla\ API, adding sensor models to the ego vehicle becomes tantamount to adding several lines in a configuration file to specify basic sensor attributes. Figure~\ref{fig:dataset-sensor-visualization} illustrates the diversity of sensor modalities present in our datasets. Our datasets generated from \carla\ are composed of RGB cameras, semantic segmentation cameras, depth cameras, \lidar, and \radar\ sensors. Our dataset generation pipeline fills a void in V2V/V2I autonomy. Infrastructure sensors will have different operational characteristics compared to vehicle-based sensing. These differences include geometry such as the location of sensors in the environment and operating modes such as computation and communication bandwidths.

\begin{figure*}[t!]
    \centering
    \includegraphics[width=0.95\linewidth]{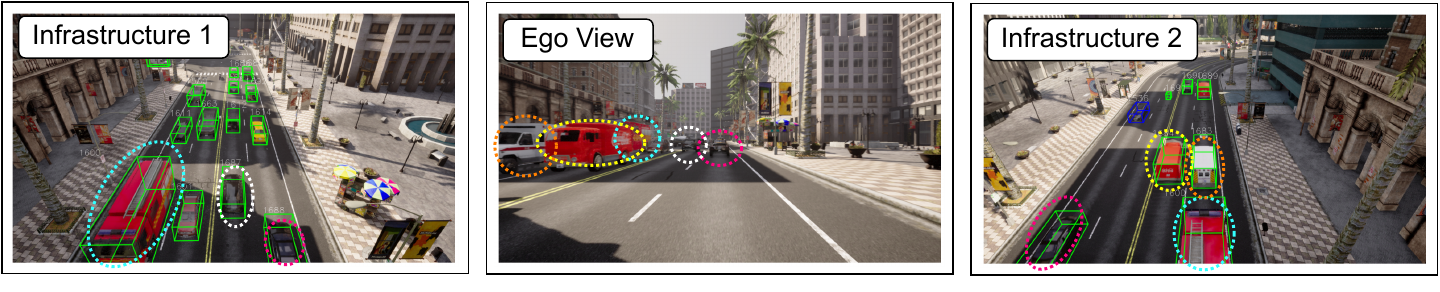}
    \caption{We provide the first dataset for infrastructure-based multi-agent collaboration. Multiple viewpoints mitigates the effects of occlusion on situational awareness. Using \avstack, we obtain consistent groundtruth labeling between multiple complementary sensors. Our novel depth-based postprocessing determines which objects are in-view for groundtruth labels. Correspondences between infrastructure and ego viewpoints shown above using manual coloring and automated object ID numbers (small font size). Ego vehicle denoted by dark blue bounding box.}
    \label{fig:multi-agent-visualization}
\end{figure*}

A selection of infrastructure data is provided in Figure~\ref{fig:multi-agent-visualization} to illustrate the benefits of multi-view geometry in perception. 

\subsection{Diverse Scenario Configuration}

In real-world datasets, it is impossible to know a-prior the scene configuration. The diversity of the scenes captured in the real-world is a product of unknown variables such as the behavior of other agents (the non-player controller characters, `NPCs'). This limits the practicality of benign real-world datasets as interesting traffic scenarios such as responding to dangerous drivers or reacting to impending crashes are not possible. 

Using a simulator engine, on the other hand, allows for custom configurations where the NPCs can be specifically triggered to perform operator-designed maneuvers. In our autonomy testing framework, the control of NPCs includes, but is not limited to, forcing lane changes at preassigned timepoints and inducing rapid changes in acceleration in the NPCs. We provide baseline scenario implementations that can be run by simply executing a shell script after starting a \carla\ docker container. 


\subsection{Multi-Agent Reference Frames}

A primary cause of the lack of development of infrastructure sensing datasets is inhomogeneous reference frames. Previous multi-agent datasets utilized a singular global coordinate frame~\cite{2022openv2v} to register perception data leveraging the fact that sensors on multiple vehicles were placed in the same configuration on each vehicle. Infrastructure sensing does not maintain that configuration assumption - both vertical placement and orientation angles will differ between platforms. The selection of \avstack\ for our dataset generation pipeline provides the benefit of \avstack's reference frame chain of command (RefChoC) that can geometrically align (i.e.,~register) perception data from arbitrary configurations in a computationally-efficient manner.

\subsection{Automated Ground Truth Labeling}

For obtaining a complete assessment of object detection and classification performance, it is essential to determine which objects are visible from the sensor. We take this for granted when using datasets captured from the real-world. A \textit{major shortcoming} of the \carla\ API is that one cannot use the client API to determine which objects are \textit{visible} to a given sensor. For example, an object may be \textit{in-view} of a camera, but using the \carla\ client API alone, it is not possible to determine if that object is \textit{visible} or e.g.~behind a building. The visibility challenge is a key reason why few datasets exist from the \carla\ simulator.

To overcome the visibility problem, we design a novel method of computing visibility from ray tracing. Using depth information from a \carla\ depth sensor, we categorize occlusion into \{\texttt{NONE}, \texttt{PARTIAL}, \texttt{MOST}, and \texttt{COMPLETE}\} categories based on the fraction of an object's bounding box within the depth image that contains a plausible depth value to its true distance. 

Algorithm~\ref{alg:occ-depth} provides the procedure for determining visibility. We use our novel method of computing visibility from depth images to filter for unoccluded objects in-view of the prescribed sensor. With knowledge of which objects are visible for each sensor, we can provide consistent groundtruth labeling between sensors.

\begin{algorithm} [!t]
\caption{Set object occlusion by depth image.}\label{alg:occ-depth}
\begin{algorithmic}
    \REQUIRE $b_{3d}$ 3D bounding box; $C$ cam. calibration; $\tau$ distance threshold (e.g.,~5); $d$ depth image to be indexed using $[\ ]$
    \ENSURE $r$ occlusion ratio on $[0,\, 1]$; occlusion value set by discretizing $r$ (e.g., $occ \gets \texttt{NONE}$ if $r \geq 0.75$).
    \vspace{4pt}
    \STATE $b_{2d} \gets b_{3d}.\text{project}(C)$ \COMMENT{project box to 2D image plane}
    \STATE $d_{b_{2d}} \gets d[b_{2d}] - \left(b_{3d}.\text{center} - \onehalf b_{3d}.\text{length}\right)$
    \STATE $r \gets \sfrac{\left(\sum |d_{b_{2d}}| \leq \tau\right)}{\text{len}(d_{b_{2d}})}$ \COMMENT{frac. of plausible depths in box}
    \RETURN r
\end{algorithmic}
\end{algorithm}


\subsection{Dataset Generation}

For model training and sensor fusion case studies, we use the dataset pipeline to create two open datasets totaling 1~TB of data: (1) multi-sensor: an ego vehicle possesses 4 RGB cameras, 4 semantic segmentation cameras, 4 depth cameras, 1 centrally-mounted LiDAR, and 1 forward-facing radar; and (2) multi-agent: an ego vehicle possesses 1 forward-facing RGB camera, 1 centrally-mounted LiDAR, and 1 forward-facing radar to go along with 5 RGB cameras, 5 LiDARs, and 5 radars placed statically (for each scene) as infrastructure sensing. Table~\ref{tab:dataset-statistics} provides a summary of our datasets compared to previous benchmarks. Note that our datasets are merely \emph{instances} of what the pipeline can do and that the pipeline can be used to generate many novel \msma\ datasets to meet the needs of the end-user.

\begin{table*}[t]
    \centering
    \caption{Comparison to benchmark multi-sensor AV datasets. Our general dataset generation pipeline has no limitation on number of sensors. For concrete analysis, we generate a multi-sensor and multi-agent dataset.}
    \label{tab:dataset-statistics}
    \begin{tabular}{c l l l l l}
        & \textbf{KITTI~\cite{2013kittidataset}} & \textbf{nuScenes~\cite{2020nuscenesdataset}} & \textbf{OPV2V~\cite{2022openv2v}} & \textbf{Ours (sensor)} & \textbf{Ours (agent)} \\
        \hline
        Sensors & 
        \tworowsubtableleft{1x LiDAR}{4x RGB Rameras} & 
        \tworowsubtableleft{1x LiDAR}{\tworowsubtableleft{6x RGB Cameras}{5x RADAR}} & 
        \tworowsubtableleft{avg. 3x CAVs}{\tworowsubtableleft{1x LiDAR per CAV}{4x RGB Camera per CAV} }& 
        \tworowsubtableleft{1x LiDAR}{\tworowsubtableleft{4x RGB Cameras}{\tworowsubtableleft{4x SemSeg Cameras}{\tworowsubtableleft{4x Depth Cameras}{1x RADAR}}}} & 
        \tworowsubtableleft{6x LiDAR}{\tworowsubtableleft{6x RGB Cameras}{6x RADAR}} 
        \\
        \hline
        Labeled Data & 
        \tworowsubtableleft{51k Objects}{\tworowsubtableleft{7481 Images}{7481 Point Clouds}} & 
        \tworowsubtableleft{ Objects}{\tworowsubtableleft{240k Images}{40k LiDAR PCs}} & 
        \tworowsubtableleft{232k Objects}{\tworowsubtableleft{40k Images}{40k LiDAR PCs}} & 
        \tworowsubtableleft{3.1M Objects}{\tworowsubtableleft{40k Images}{\tworowsubtableleft{20k LiDAR PCs}{20k Radar PCs}}} & 
        \tworowsubtableleft{1M Objects}{\tworowsubtableleft{30k Images}{\tworowsubtableleft{30k LiDAR PCs}{30k Radar PCs}}} 
        \\
        \hline
        \hline
        Dataset Volume & 40~GB & 280~GB & 250~GB & 500~GB & 500~GB \\
        Labeled Data Volume & 20~GB & 280~GB & 250~GB & 500~GB & 500~GB \\
        \hline
    \end{tabular}
    \vspace{-10pt}
\end{table*}

\section{Supervised Perception Models} \label{sec:3-models}

Beyond the generation of the datasets, we provide a framework for training dozens of domain \textit{and} application-specific perception models. Integration with OpenMMLab's perception algorithms enables dozens of 2D object detection, 3D object detection, and 2D semantic segmentation models. Training models on scenes generated from \carla\ was previously untenable due to the glaring lack of consistent groundtruth labeling from any saved \carla\ data.

\vspace{4pt}
\noindent\textbf{Application-Specific Models.}
%
As industry moves closer to deploying solutions for \msma\ autonomy, in-depth research becomes paramount. Importantly, ground, aerial, and infrastructure sensors operate under different data distributions. Without a suitable labeled dataset for each perception applications, general-purpose models must be used and will underperform.

\begin{figure}
    \centering
    \begin{subfigure}{0.45\linewidth}
        \fbox{\includegraphics[trim={2.5cm 1cm 3.5cm 1cm},clip,width=0.9\linewidth]{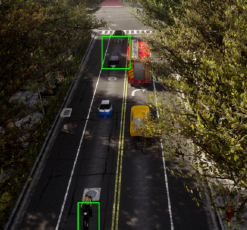}}
        \caption{Trained on ego}
    \end{subfigure}
    \begin{subfigure}{0.45\linewidth}
        \fbox{\includegraphics[trim={2.5cm 1cm 3.5cm 1cm},clip,width=0.9\linewidth]{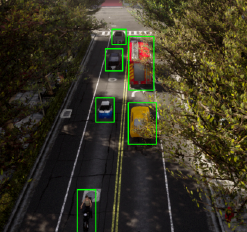}}
        \caption{Trained on infrastructure}
    \end{subfigure}
    \caption{Perception algorithms must be trained on application-specific datasets. When testing on infrastructure data, (a) training on the ego vehicle's perception data is inferior to (b) training on representative data from infrastructure sensing.}
    \label{fig:application-need}
\end{figure}

To validate these concerns, we provide an example in Figure~\ref{fig:application-need} where a camera-based model trained from only ground-vehicle application (e.g.,~KITTI, nuScenes, OPV2V data) is used to detect objects in an infrastructure application. This model fails to adequately detect target objects because it was not presented with this application during tarining. The lack of success motivates training models on application-relevant datasets. Until now, this result was not feasible due to a lack of available infrastructure and multi-agent datasets. 

We leverage our generated datasets to train 2D and 3D object detection and classification models for vehicle-based and infrastructure-based applications. In a detailed analysis of perception performance, we consider three scenarios for each sensor class: (1) training and testing on vehicle, (2) training on vehicle, testing on infrastructure, and (3) training and testing on infrastructure. We run each case over the validation set of our datasets and capture the class-wise average precision of the results. 

\begin{figure}
    \centering
    \begin{subfigure}{0.95\linewidth}
        \centering
        \includegraphics[width=0.88\linewidth]{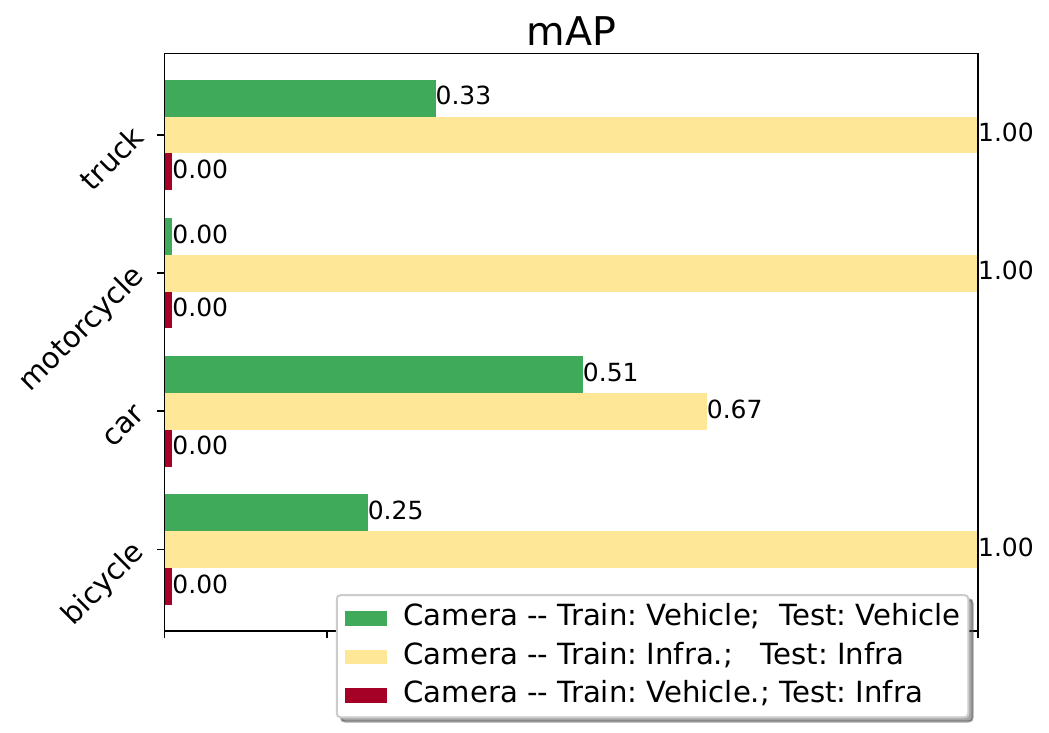}
        \caption{mAP outcomes on camera perception (validation).}
    \end{subfigure}
    \begin{subfigure}{0.95\linewidth}
        \centering
        \includegraphics[width=0.88\linewidth]{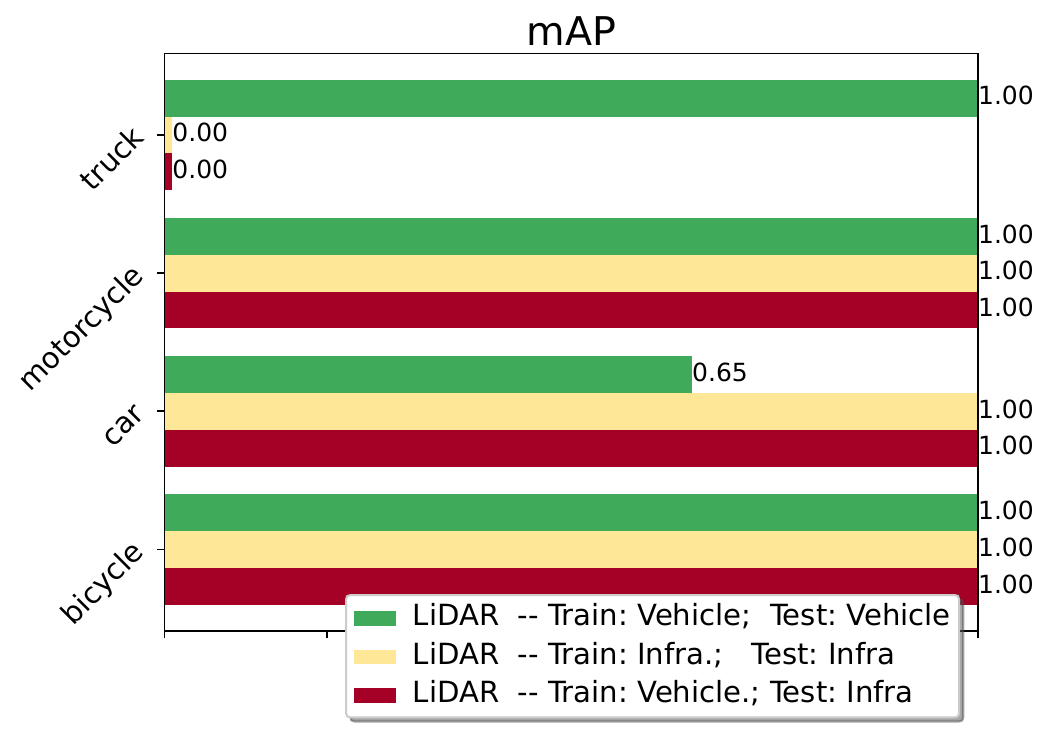}
        \caption{mAP outcomes on LiDAR perception (validation).}
    \end{subfigure}
    \caption{Our framework for multi-agent autonomy enables training application-specific models. (a) Camera models trained on vehicle data (green) cannot perform adequately in the infrastructure application (yellow). Vantage point of infrastructure sensor (yellow) performs better than in vehicle (red), suggesting strong benefit of using infrastructure collaboration. (b) \lidar\ models perform nearly equivalently when training/testing in vehicle and infrastructure applications. Results consistent with the point cloud as a viewpoint-agnostic 3D scene representation. Perception performance indicates that using application-specific models substantially outperforms reuse of monolithic perception models.}
    \label{fig:application-results}
\end{figure}

Outcomes of this experiment are illustrated in Figure~\ref{fig:application-results}. Clearly, in the case of the camera sensor, training on vehicle data and testing on infrastructure data completely fails to detect objects. On the other hand, this is less of a problem in the LiDAR data case. This is because LiDAR is a full 3D pointcloud which is only minimally impacted by sensor translation and rotation.

\section{Multi-Agent Case Studies} \label{sec:4-algorithms}

The AV designer must decide the level at which data fusion occurs in multi-agent pipelines. Two attractive options for the level of data fusion include at \textit{perception} and \textit{tracking}. However, in prior analysis, the researchers placed large assumptions on the \emph{purity} of the sensor network data. In particular, multi-agent network topologies in~\cite{2022openv2v} were designed such that a central agent received all data from nearby agents and that those agents solely relied on their own information. We describe why this is an unrealistic assumption and use our framework to model the outcomes of an unknown sensor network. To handle such challenges, we introduce fusion at the \textit{post-tracking} level in the form of a dedicated data fusion module.

\subsection{Multi-Agent Fusion Algorithms}
\begin{figure}
    \centering
    \begin{subfigure}[b]{0.49\linewidth}
        \includegraphics[width=\linewidth]{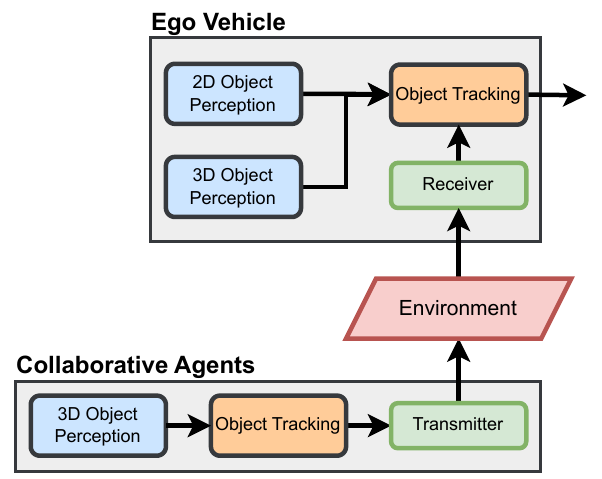}
        \caption{Fusion at tracking.}
    \end{subfigure}
    \begin{subfigure}[b]{0.49\linewidth}
        \includegraphics[width=\linewidth]{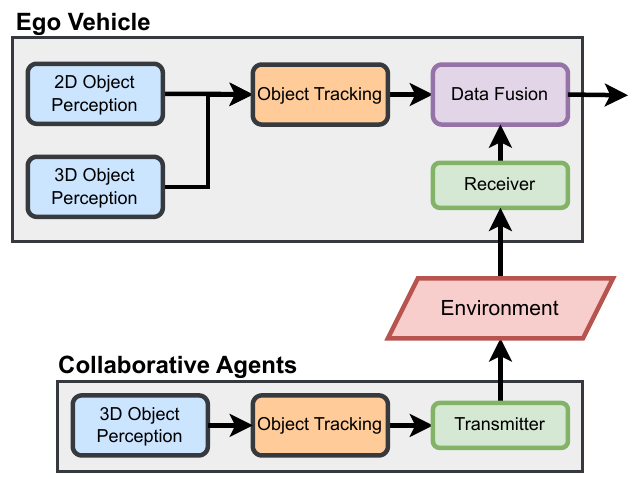}
        \caption{Dedicated fusion.}
    \end{subfigure}
    \caption{Data fusion can occur at multiple points in the pipeline. (a) Fusion at tracking is optimal when temporal and inter-sensor correlations are known. Otherwise, (b) dedicated fusion models that conservatively consider unknown correlations can lead to more consistent situational awareness.}
    \label{fig:collaborative-algorithms}
\end{figure}

First, we present the merits and drawbacks of multi-agent data fusion presented in several prior works. An illustration of the fusion algorithms and multi-agent network topologies is provided to accompany this section in Figure~\ref{fig:collaborative-algorithms} and results of the experiments are presented in Figure~\ref{fig:collaborative-results}.

\subsubsection{Fusion at Perception}
From the perspective of maximizing sharing of available information, one option for data fusion is to share the raw perception data (or low-level features) between agents. This was an approach taken in OPV2V~\cite{2022openv2v} and can obtain accurate results due to the richness of the shared data. Unfortunately, sharing low-level data is not always possible due to limited wireless bandwidth. Sharing large volumes of data will introduce latencies into the fusion pipeline. Moreover, the integrity of raw data is difficult to verify (e.g.,~consider imperceptible adversarial examples~\cite{carlini2017adversarial}), leaving low-level fusion vulnerable to adversarial manipulations. Such limitations make sharing low-level sensor data unlikely to be realized in practice~\cite{nhtsav2v}.

\subsubsection{Fusion at Tracking}
If transmitting raw data is infeasible, the next candidate is fusion is at tracking. In this case, perception is performed on each agent's data, individually. Then, the output of perception as high-level semantic information (e.g.,~object detections) is passed between agents. Upon receipt, each agent treats the incoming information as a new raw detection to update its own situational awareness. This ``detection-level'' fusion is tantamount to treating each sensor independent and sharing detections as if they were uncorrelated measurements. \cite{2022openv2v} assumes this correlation assumption holds. 

In the real world, uncorrelated detection-level fusion is infeasible. This is because detections are noisy and must be longitudinally filtered to remove false-positives and estimate higher-order parameters such as velocity. It is more likely that each platform will perform both perception \emph{and} tracking. This will induce temporal correlations in data from each platform. Moreover, each agent will be receiving information from other platforms who in-turn will have received information from \emph{others}. The sensor network will not maintain a ``clean'' structure, in contrast to the naive modeling of small case studies (e.g.,~\cite{2022openv2v}). Temporal and inter-sensor correlations due to complicated network topologies make it unwise to share information at tracking. Cycles and correlations can lead to statistical overconfidence if the correlations are not accurately modeled~\cite{2017ddfwithCI, 1994ddfframework}. 

\subsubsection{Fusion Post-Tracking: DDF}

Recent results suggest the AV community is considering fusion as a \textit{centralized} problem. Case studies tend to model agents sending purely-local information to the ego agent (e.g., in~\cite{2022openv2v, 2022cooperative}). However, in the real-world, collaborative agents cannot be assumed to only be processing local information. Other agents will be sending and receiving data in a complex sensor network. As such, it is more accurate to consider the problem of multi-agent autonomy as a \textit{decentralized sensor network} problem.

\setlength{\fboxsep}{2pt}
\begin{figure*}[t!]
    \centering
    \begin{subfigure}[b]{0.24\linewidth}
        \begin{tikzpicture}
            \node(a){\fbox{
            \includegraphics[trim={8cm 8cm 2.5cm 5cm},clip,width=0.92\linewidth]{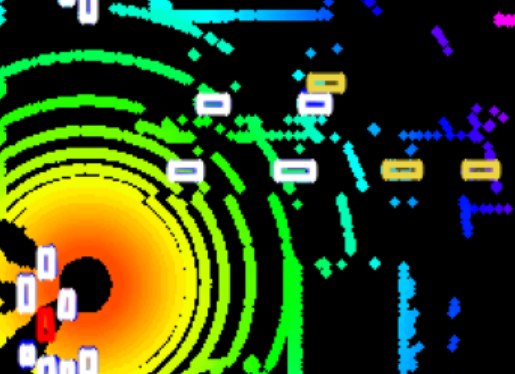}
            }};
            \node at(a.center)[draw, cyan,line width=2pt,ellipse, minimum width=40pt, minimum height=40pt,rotate=0,xshift=5pt,yshift=10pt]{};
        \end{tikzpicture}
        \caption{Outcomes: ego local.}
    \end{subfigure}
    \begin{subfigure}[b]{0.24\linewidth}
        \begin{tikzpicture}
            \node(a){\fbox{
            \includegraphics[trim={6cm 0cm 0cm 0cm},clip,width=0.885\linewidth]{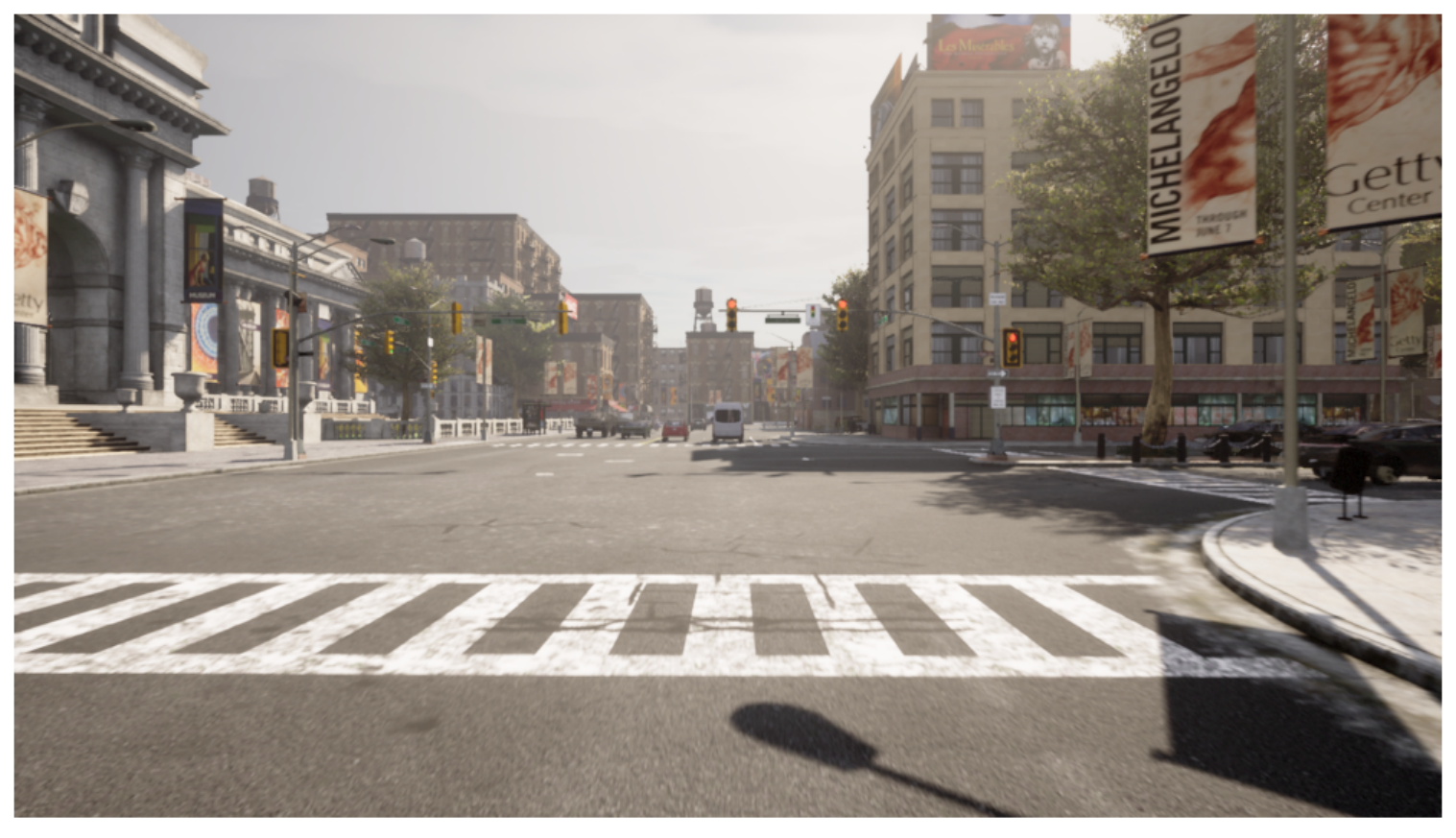}
            }};
            \node at(a.center)[draw, cyan,line width=2pt,ellipse, minimum width=50pt, minimum height=60pt,rotate=0,xshift=35pt,yshift=5pt]{};
        \end{tikzpicture}
        \caption{Forward view: ego}
    \end{subfigure}
    \begin{subfigure}[b]{0.24\linewidth}
        \begin{tikzpicture}
            \node(a){\fbox{
            \includegraphics[trim={8cm 8cm 2.5cm 5cm},clip,width=0.92\linewidth]{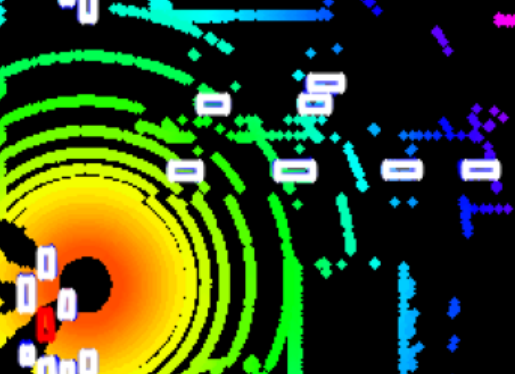}
            }};
            \node at(a.center)[draw, white,line width=2pt,ellipse, minimum width=55pt, minimum height=40pt,rotate=0,xshift=25pt,yshift=10pt]{};
        \end{tikzpicture}
        \caption{Outcomes: ego + infra}
    \end{subfigure}
    \begin{subfigure}[b]{0.24\linewidth}
        \begin{tikzpicture}
            \node(a){\fbox{
            \includegraphics[trim={4cm 0cm 2cm 0cm},clip,width=0.885\linewidth]{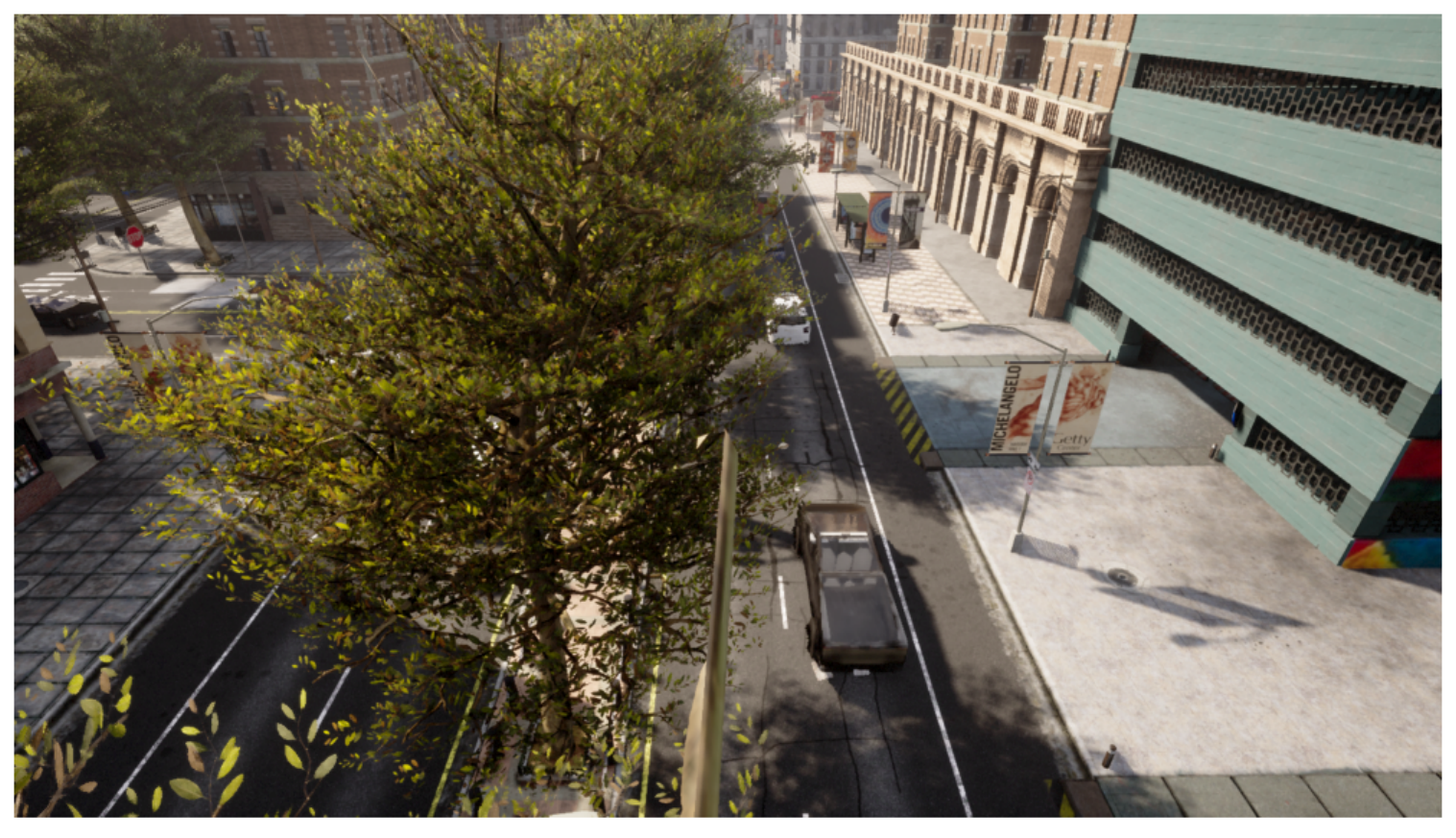}
            }};
            \node at(a.center)[draw, white,line width=2pt,ellipse, minimum width=40pt, minimum height=70pt,rotate=0,xshift=0pt,yshift=-5pt]{};
        \end{tikzpicture}
        \caption{Forward view: infrastructure}
    \end{subfigure}
    \caption{In birds eye view point cloud (a,c), white/blue are correct track assignments, yellow are false negatives, red are false positives. (a) Presence of \emph{false negatives} indicates that ego alone is not capable of complete situational awareness. This is due to (b) occlusions in ego's field of view. (c) When incorporating infrastructure sensing, false negatives are eliminated without any change in false positives (red). This is due to (d) infrastructure's unique vantage points that can mitigate occlusions.}
    \label{fig:collaborative-visualization}
\end{figure*}
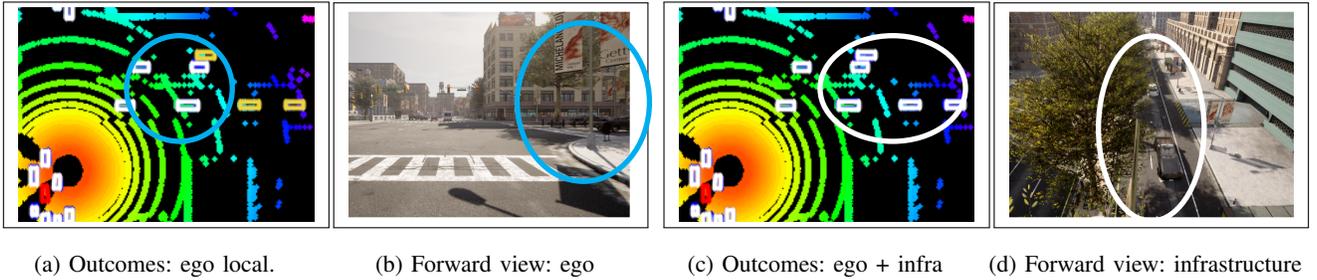

To handle correlated networks, we introduce fusion at the post-tracking level leveraging the rich history of distributed data fusion~\cite{1994ddfframework}. Each agent begins by locally maintaining its own tracks that derive from its local perception data. Once agents begin to share tracked object information with neighboring agents, tracks are fused between agents by first associating overlaps in the two sets of tracks with a classical assignment (e.g.,~JVC~\cite{1986JVC}) and then by merging assignments with a conservative data fusion algorithm such as covariance intersection~\cite{2017ddfwithCI}. Prior multi-agent case studies that we have observed (e.g.,~\cite{2022openv2v, 2022cooperative}) could not consider algorithms such as this because they did not appropriately model the very-likely possibility of a network of sensors that have time-correlated and platform-correlated information.

\subsection{Multi-Agent Experiment}

To test the fusion configurations, we design an experiment that models the ego agent's data fusion system and a multi-agent (i.e.,~infrastructure-aided) network topology using the \emph{multi-agent} dataset generated using our proposed framework. We consider three agent models:
\begin{itemize}
    \item \textbf{Local:} Ego only uses local perception data.
    \item \textbf{Fusion at tracking:} Ego fuses data shared from other agents in the tracking module.
    \item \textbf{Fusion post-tracking:} Ego fuses data shared from other agents with its own tracks using distributed data fusion.
\end{itemize}
We also consider three network topology models:
\begin{itemize}
    \item \textbf{No correlation:} Each infrastructure sensor uses only local perception data.
    \item \textbf{Minor correlation:} Each infrastructure sensor runs perception and tracking on local data. With some small probability, each agent will send its detection information to the other infrastructure agents.
    \item \textbf{Major correlation:} Same as \textbf{minor correlation} but with high probability of sending information.
\end{itemize}

\begin{figure}[t!]
    \centering
    \vspace{-10pt}
    \includegraphics[width=0.8\linewidth]{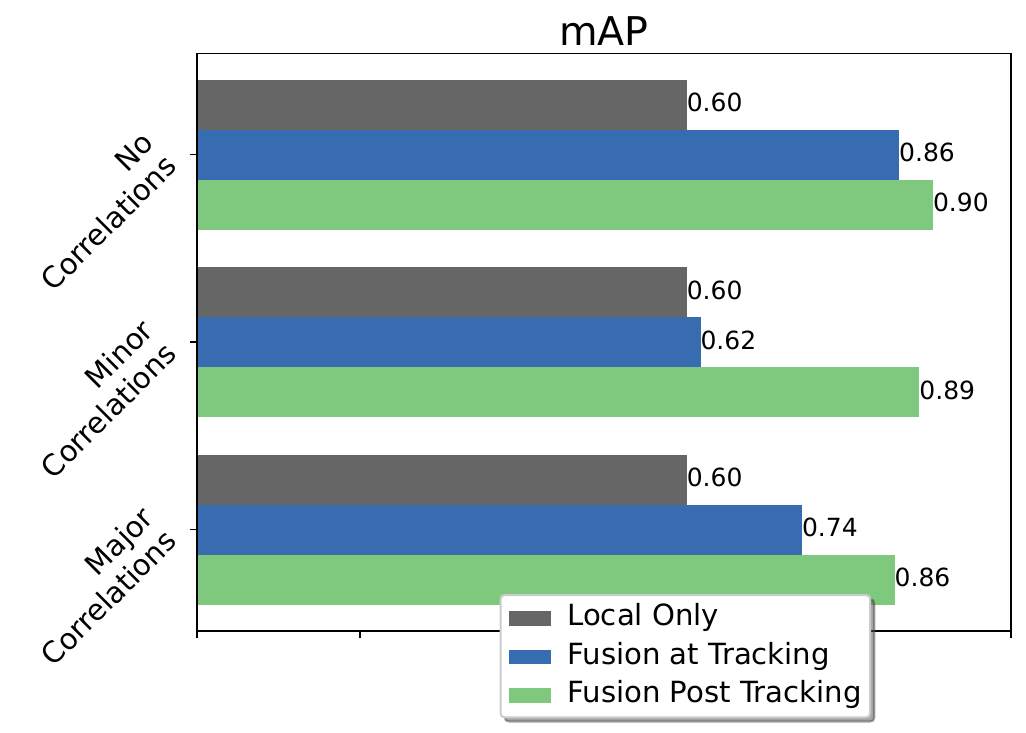}
    \caption{Ego-infrastructure collaboration improves mAP outcome in all network topologies. In cases of intra-network correlations (minor/major), detection-level performance deteriorates compared to track-to-track fusion. Existing algorithms for naive multi-agent fusion will underperform and more research is needed.}
    \label{fig:collaborative-results}
\end{figure}

We run each pair of ego and network combination over the validation scenes from our multi-agent dataset. After a burn-in time period, we compare outcomes to the groundtruth object locations within 75~m of the ego agent. To assess the performance, we calculate the average precision (AP) aggregated over all the classes. The results are included in Figure~\ref{fig:collaborative-results}.
. As expected, in the case of no correlations in the sensor data, both fusion approaches outperform when the ego agent uses only local information. This outcome is illustrated in Figure~\ref{fig:collaborative-visualization} and explained by that infrastructure sensors can mitigate strong occlusion relationships that otherwise deteriorate an agents situational awareness.

Importantly, we also find that, in the case of minor and major correlations in the data shared amongst the agents in the network, the post-tracking fusion with the DDF algorithm outperforms treating incoming tracks as raw measurements by fusion at the tracking level. This outcome is consistent with the intuition that fusing correlated tracks at the tracking level can lead to overconfidence about false positives which decreases the mAP score.

\section{Conclusion}

We  introduced a framework for datasets, models, and autonomy case studies for multi-sensor, multi-agent environments. Our framework is built on \avstack\ and leverages integration with the \carla\ simulator. Using this new framework, we  developed novel multi-agent datasets and considered for the first time infrastructure sensing. We then used these datasets to train application-specific perception models and illustrate the utility of these models in pursuing collaborative algorithms in complex sensor networks.

\FloatBarrier

\bibliographystyle{IEEEtran}
\bibliography{references}


\end{document}